\documentclass[runningheads]{llncs}

\usepackage{eccv}

\usepackage{eccvabbrv}

\usepackage{graphicx}
\usepackage{booktabs}

\usepackage[accsupp]{axessibility}  %

\usepackage{hyperref}

\usepackage{orcidlink}

\usepackage{amsmath}
\usepackage{bbm}
\usepackage{multirow}
\usepackage{xcolor}
\DeclareMathOperator*{\argmax}{arg\,max}

\newcommand{\figref}[1]{Fig.~\ref{#1}}
\newcommand{\tabref}[1]{Tab.~\ref{#1}}
\newcommand{\secref}[1]{Sec.~\ref{#1}}

\newcommand{\equref}[1]{Eqn.~(\ref{#1})}
\usepackage{colortbl}

\newcommand{\colwidthA}{1.0cm}

\newcolumntype{L}[1]{>{\raggedright\let\newline\\\arraybackslash\hspace{0pt}}m{#1}}
\newcolumntype{C}[1]{>{\centering\let\newline\\\arraybackslash\hspace{0pt}}m{#1}}
\newcolumntype{R}[1]{>{\raggedleft\let\newline\\\arraybackslash\hspace{0pt}}m{#1}}

\begin{document}

\title{IG Captioner: Information Gain Captioners \\ are Strong Zero-shot Classifiers} 

\titlerunning{Information Gain Captioner}

\author{Chenglin Yang\inst{1} \and
Siyuan Qiao\inst{2} \and
Yuan Cao\inst{2} \and
Yu Zhang\inst{2} \and \\
Tao Zhu\inst{2} \and
Alan Yuille\inst{1} \and
Jiahui Yu\inst{2}
}

\authorrunning{C. Yang et al.}

\institute{Johns Hopkins University \and
Google DeepMind
}

\maketitle

\begin{center}
 \centering
 \small
 \setlength{\tabcolsep}{0.0pt}
 \begin{tabular}{c}
    \includegraphics[width=1.0\textwidth]{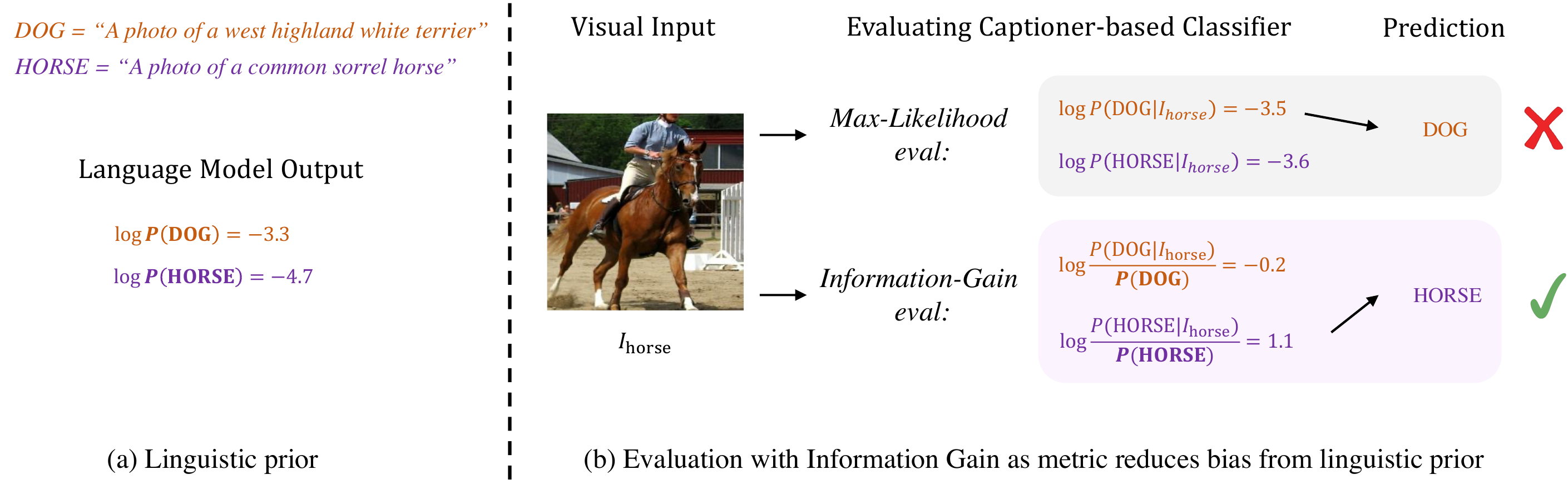}
 \end{tabular}
 \captionof{figure}{The prediction of a captioner-based classifier is influenced by the linguistic priors in pure text modality. The Information Gain (IG) evaluation reduces such impact and makes the predictions more grounded on the visual inputs. We illustrate with \textbf{real} predictions on zero-shot ImageNet~\cite{deng2009imagenet} classification in this figure. (a) The language model was trained on Laion-5B~\cite{schuhmann2022laion} captions only. (b) The captioner was trained on the Laion-5B dataset with both the images and captions. The IG evaluation in (b) uses the outputs of both the captioner and the language model in (a).
 }
 \label{fig: bias_from_linguistic_prior}
\end{center}

\begin{abstract}

Generative training has been demonstrated to be powerful for building visual-language models. However, on zero-shot discriminative benchmarks, there is still a performance gap between models trained with generative and discriminative objectives. In this paper, we aim to narrow this gap by improving the efficacy of generative training on classification tasks, without any finetuning processes or additional modules.

Specifically, we focus on narrowing the gap between the generative captioner and the CLIP classifier. We begin by analysing the predictions made by the captioner and classifier and observe that the caption generation inherits the distribution bias from the language model trained with pure text modality, making it less grounded on the visual signal. To tackle this problem, we redesign the scoring objective for the captioner to alleviate the distributional bias and focus on measuring the gain of information brought by the visual inputs. We further design a generative training objective to match the evaluation objective. We name our model trained and evaluated from the novel procedures as Information Gain (IG) captioner. We pretrain the models on the public Laion-5B dataset and perform a series of discriminative evaluations. For the zero-shot classification on ImageNet, IG captioner achieves $> 18\%$ improvements over the standard captioner, achieving comparable performances with the CLIP classifier. IG captioner also demonstrated strong performance on zero-shot image-text retrieval tasks on MSCOCO and Flickr30K.
We hope this paper inspires further research towards unifying generative and discriminative training procedures for visual-language models.

\end{abstract}

\section{Introduction}

Generative training has been shown to be highly effective for both the language~\cite{brown2020language} and vision models~\cite{sohl2015deep, ho2020denoising}. These generative models demonstrate remarkable zero-shot generation capabilities and have led to the developments of many impressive AI applications~\cite{OpenAI_GPT4_2023, rombach2022high}. On the other hand, discriminative training is still an effective procedure for building the large vision-language models like CLIP~\cite{radford2021learning} and CoCa~\cite{yu2022coca}. Models trained with discrminative objectives naturally performs strongly on discriminative tasks like classification.

The inconsistent setups between generative and discriminative training raises an intriguing topic, which we aim to address in this paper: \textit{Can generative training procedure also perform well on zero-shot discriminative tasks}?
Concretely, 
we study the performance gap between generative captioners and discriminative models on zero-shot classification tasks, and propose novel approaches to narrow the gap.

To obtain an effective zero-shot generative classifier, we start with evaluating a standard generative captioner with the Maximum Likelihood Estimation (MLE) objective on classification tasks. Specifically, we pretrain a captioner on the public Laion-5B~\cite{schuhmann2022laion} dataset and evaluate it on the ImageNet~\cite{deng2009imagenet}. We also pretrain a CLIP classifier with the same experimental setting.
We observe a large performance gap between the captioner and the CLIP classifier. 
We find that the predictions of the captioner are less grounded on the visual inputs. Instead, they are
biased by the linguistic priors, which is illustrated in~\figref{fig: bias_from_linguistic_prior}.
These observations indicate that we should discount the influence by predictions from pure text modality in multimodal generative classification setup.

To deal with this problem, we propose an Information Gain (IG) evaluation objective, as depicted in~\figref{fig: inference}. This objective demonstrates a significant performance boost on top of a captioner trained with conventional procedure.
Furthermore, we propose a generative training objective to match the IG evaluation. We name the captioner, trained and evaluated with our proposed objectives, as Information Gain (IG) captioner.
We pretrain the models on the Laion-5B dataset and evaluate multiple zero-shot discriminative tasks. All the models have the same training and evaluation settings to ensure the fair comparisons. For zero-shot ImageNet classification, IG captioner with ViT-B/ViT-L~\cite{dosovitskiy2020image} image encoder shows 19.7\%/18.1\% improvements on top-1 accuracy over the standard captioner. It is even better than the CLIP classifier by 0.5\%/1.6\% top-1 accuracy. For zero-shot image-text retrieval on MSCOCO~\cite{chen2015microsoft} dataset, IG captioner demonstrates $>$ 22\% accuracy improvements on all of image-to-text R@1, R@5 and R@10 over the captioner, significantly reducing the gap from the captioner to the CLIP classifier. Interestingly, we observe the superior performances on text-to-image recalls of both the captioner and IG captioner over the CLIP classifier. The superiority of IG captioner is also demonstrated on Flickr30K~\cite{plummer2015flickr30k} dataset.

\renewcommand{\colwidthA}{2.0cm}
\begin{table}[!t]
 \centering
 \small
 \setlength{\tabcolsep}{0.0pt}
 \begin{tabular}{c}
    \includegraphics[width=1.0\textwidth]{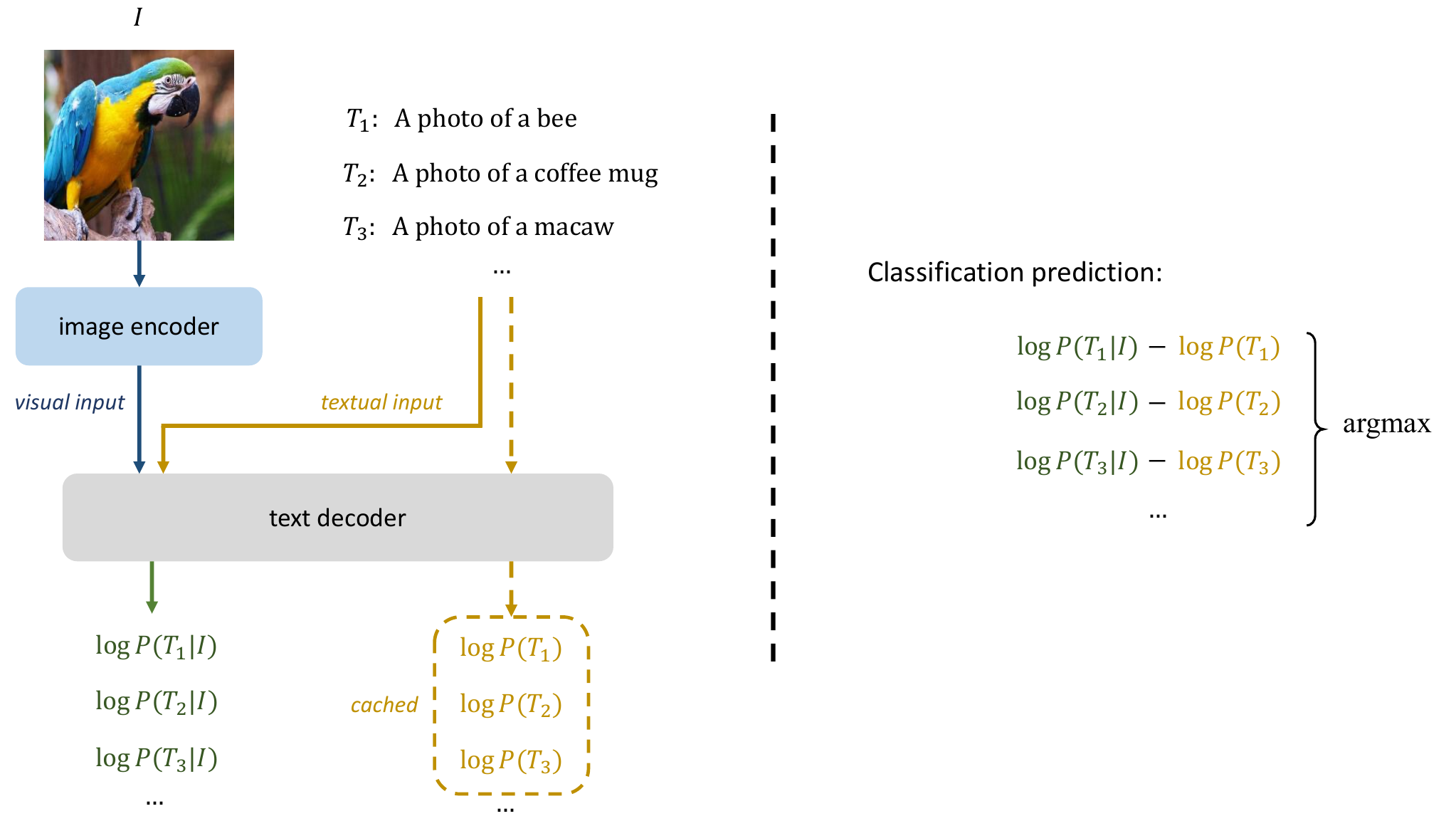}
 \end{tabular}
 \captionof{figure}{The inference pipeline of IG captioner. $I$ and $T$ represent the input image and caption. IG captioner consists of a image encoder and a text decoder. The text decoder is able to provide both the multimodal and unimodal predictions. The unimodal predictions can be cached for different input images. Both the image and object classes are from the ImageNet dataset.}
 \label{fig: inference}
\end{table}

Our main contributions can be summarized in the following:
\begin{itemize}
    \item We convert a generative captioner into a zero-shot classifier. We identify the root cause of its poor performance and observe that its predictions are negatively influenced by the linguistic priors and less grounded on the visual inputs. 
    \item We propose the Information Gain (IG) method to evaluate a generative classifier. We demonstrate that the IG evaluation is able to significantly improve its performance without any changes in the trainng process.
    \item To couple with the IG evaluation, we propose a generative training objective to further boost the performance of the generative classifier which we name Information Gain (IG) captioner.
    \item We show that IG captioner is a strong zero-shot classifier, evidenced by the significant improvements over the captioner and comparable performance with the CLIP classifier on the zero-shot ImageNet classification, MSCOCO and Flickr30K image-text retrieval tasks.
\end{itemize}

\section{Related Work}

\textbf{\quad Text-to-image Generative Classifiers.}
The classic approach~\cite{hinton2007recognize,ranzato2011deep} of using generative models to perform recognition tasks is the Bayes algorithm~\cite{ng2001discriminative}. During training, this algorithm models the data distribution while during inference, it provides predictions by solving a maximum likelihood estimation (MLE) problem. Recently, there have been the methods~\cite{li2023your,clark2023text} proposed to convert the text-to-image diffusion models, Stable Diffusion~\cite{rombach2022high} or Imagen~\cite{saharia2022palette}, into a zero-shot classifier using the Bayes algorithm. 

The focus of IG captioner is different with these previous arts. First, we explore the possibility of achieving a good classifier solely through the generative training, instead of how to convert a generative model into a classifier. The Stable Diffusion used by the above diffusion model classifiers~\cite{li2023your,clark2023text} is not suitable for our goal because it uses the CLIP text encoder pretrained with contrastive loss to provide the text guidance. Second, we identify the negative impact of the text priors inherited from the pretraining data and propose methods to reduce this impact for zero-shot classification tasks. Besides, IG captioner is an image-to-text generative captioner. We also perform the comparisons in~\tabref{tab: compare_image_to_text}.

\textbf{Classifier-Free Guidance (CFG).}
CFG~\cite{ho2022classifier} is an important approach to improving the sample quality of text-to-image generative models~\cite{nichol2021glide,ramesh2022hierarchical,saharia2022photorealistic,gafni2022make,yu2022scaling}. During training, CFG requires the model to generate images both with and without text conditions. During inference, the output image is sampled according to a linear combination of both the conditional and unconditional predictions to improve text-image alignment, which relates CFG to IG captioner from a high level. In contrast, IG captioner is an image-to-text captioner that uses the probability gain from unconditionally to conditionally generating captions with the aim of reducing the bias from text priors for zero-shot classification tasks.

\textbf{Contrast in Text Generation.}
In the field of text generation, researchers have been investigating the use of contrast, including avoiding the text degeneration and undesirable repetitions~\cite{su2022contrastive}, generating pan sentences~\cite{he2019pun}, and language detoxification~\cite{liu2021dexperts}. Contrastive decoding~\cite{li2022contrastive} uses the log-likelihood difference between a large expert and a small amateur model to improve the quality of the open-ended text generation. Li et al.~\cite{li2015diversity} propose to use the mutual information between an input message and its responses as the decoding objective to avoid the generic and meaningless conversations. Differently, in this paper, IG captioner uses the Information Gain (IG) to address the problem of evaluating a vision-language generative model as a zero-shot classifier. Importantly, the IG evaluation is used to mitigate the negative influence of linguistic priors inherited from the pretraining dataset.

\textbf{Image-to-text Multimodal Language Models.}
Multimodal language models have become the dominant approaches for image captioning, including OFA~\cite{wang2022ofa}, Flamingo~\cite{alayrac2022flamingo}, BLIP~\cite{li2023blip,li2022blip}, CoCa~\cite{yu2022coca}, LLaVA~\cite{liu2024visual,liu2023improvedllava}, MiniGPT-4~\cite{zhu2023minigpt}, InstructBLIP~\cite{instructblip}, Kosmos~\cite{kosmos-1,peng2023kosmos}, Qwen-VL~\cite{bai2023qwen}, VFC~\cite{ge2024visual}, LLaVA-NeXT~\cite{liu2024llavanext,li2024llavanext-strong}, Gemini~\cite{team2023gemini,reid2024gemini}, and GPT-4~\cite{achiam2023gpt}. Differently, IG captioner demonstrates the potential of pure generative training on discriminative tasks using public datasets.

\section{Method}

In this section, we introduce Information Gain (IG) captioner. First, we illustrate how to evaluate a conventional captioner on the zero-shot classification task and diagnose its problem. Second, we propose an evaluation approach using the information gain to tackle the problem. Third, we design a training objective to couple with the IG evaluation. Combining the above evaluation and training procedures, we propose IG captioner.

\subsection{Evaluating Captioners on Discriminative Tasks}
A captioner is a generative model that generates caption for a given image. It is by nature not trained as a classifier and its output can not be used directly for such tasks like image classification or image-text retrieval. Therefore, to evaluate a captioner on discriminative tasks, we need to adapt the evaluation procedure.

\subsubsection{Maximum Likelihood Estimation (MLE) Objective.} Let $I \in \mathbb{R}^{H \times W \times 3}$, $T \in \mathbb{R}^{N \times D}$ and $t \in \mathbb{R}^{D}$ be the image, caption and the word in the caption, where $H$, $W$ are the height and width of the image and $N$, $D$ are the number of text tokens and the token dimension. A captioner is a function $F_\theta: I \rightarrow T$ where $\theta$ is the model parameter.

Discriminative tasks require the model to pick the most appropriate text from a set of candidates $C=\{T_i\}$ to describe an image, where $i$ is the index of captions. During inference, the standard MLE evaluation objective is:
\begin{equation}
    \argmax_{i} \left[ \log {P(T_{i}|I)} \right]
\label{equ: standard_eval}
\end{equation}
We omit the normalization factor $1 / N_{i}$ when calculating the evaluation objectives throughout this paper for simplicity.
$N_{i}$ is the token length of the $t^{\text{th}}$ caption.

\subsubsection{Low Performance Using MLE Objective.} To study the effectiveness of MLE as objective, we pretrain a captioner on Laion-5B and use it to perform zero-shot ImageNet classification.
We set up a strong baseline: a CLIP classifier which is pretrained with contrastive loss and tailored for classification tasks. The captioner has a image encoder and a text decoder while the CLIP classifier has a image encoder and a text encoder.
Both of them have the same model size, training and evaluation settings to ensure fair comparisons.
The results are shown in~\tabref{tab: standard_eval}. When directly using the MLE objective (\equref{equ: standard_eval}) the captioner shows $>19\%$ top-1 accuracy degradation compared with the CLIP classifier.

\begin{table}[t]
  \centering
  \begin{tabular}{cc|c}
    \toprule
    models & pretrain loss & zero-shot IN acc. (\%) \\
    \midrule
    CLIP classifier & contrast & 63.8 \\
    captioner & generate & 44.6 \\
    \bottomrule
  \end{tabular}
  \caption{The large performance gap between the standard captioner and the CLIP classifier on zero-shot ImageNet classification. All the models are trained with the Laion-5B dataset. The CLIP classifier is trained with contrastive loss while the captioner with generative loss. The CLIP classifier has a 12-layer image encoder and a 12-layer text encoder and the captioner has a 12-layer image encoder and a 12-layer text decoder.}
  \label{tab: standard_eval}
\end{table}

\begin{table}[t]
  \centering
  \begin{tabular}{cc|c}
    \toprule
    variable 1 & variable 2 & $r_{\text{var1}, \text{var2}}$  \\
    \midrule
    \multirow{1}{*}{$\log P(T)$} & \multirow{1}{*}{$\log P(T|I)$} & \multirow{1}{*}{0.77} \\
    \multirow{1}{*}{$\log P(T)$} & \multirow{1}{*}{$ \log P(T|I) - \log P(T)$} & \multirow{1}{*}{-0.39} \\
    \bottomrule
  \end{tabular}
  \caption{Mean Pearson Correlation Coefficient (PCC) measurements of zero-shot predictions on ImageNet. $\log P(T)$ is predicted by the unimodal language model which is trained with the Laion-5B captions. $\log P(T|I)$ is predicted by the standard captioner trained with both the Laion-5B images and captions. Given one image, there are 80K (1K classes, 80 prompts) observations of these two variables to compute the PCC. The mean PCC is obtained by averaging the PCC on 50K ImageNet validation images.}
  \label{tab: pcc_pti_pt}
\end{table}

\subsubsection{Result Analaysis of MLE Evaluation.} Since the captioner is a vision-language model, we are interested in understanding how much visual information is provided by the visual inputs and utilized by the model. We pretrain a language model that has the same architecture as the text decoder of the captioner.
It models the captions in the Laion-5B dataset and is able to predict $\log P(T)$. There are 1K classes on ImageNet and for each class we use the 80 prompts proposed by CLIP. In total, there are 80K unique captions. Given an image $I$, we consider $\log P(T|I)$ and $\log P(T)$ as two variables and compute their Pearson correlation coefficient (PCC)~\cite{pearson1895vii,te2015basic} with 80K observations in~~\tabref{tab: pcc_pti_pt}. 
We observe a strong correlation between $\log P(T|I)$ and $\log P(T)$, which indicates that the captioner is strongly biased by the text priors and tends to ignore the visual information. The correlation is also visualized in~\figref{fig: corr}.

\renewcommand{\colwidthA}{2.0cm}
\begin{table*}[!htp]
 \centering
 \setlength{\tabcolsep}{0.0pt}
 \begin{tabular}{c}
    \includegraphics[width=1.0\textwidth]{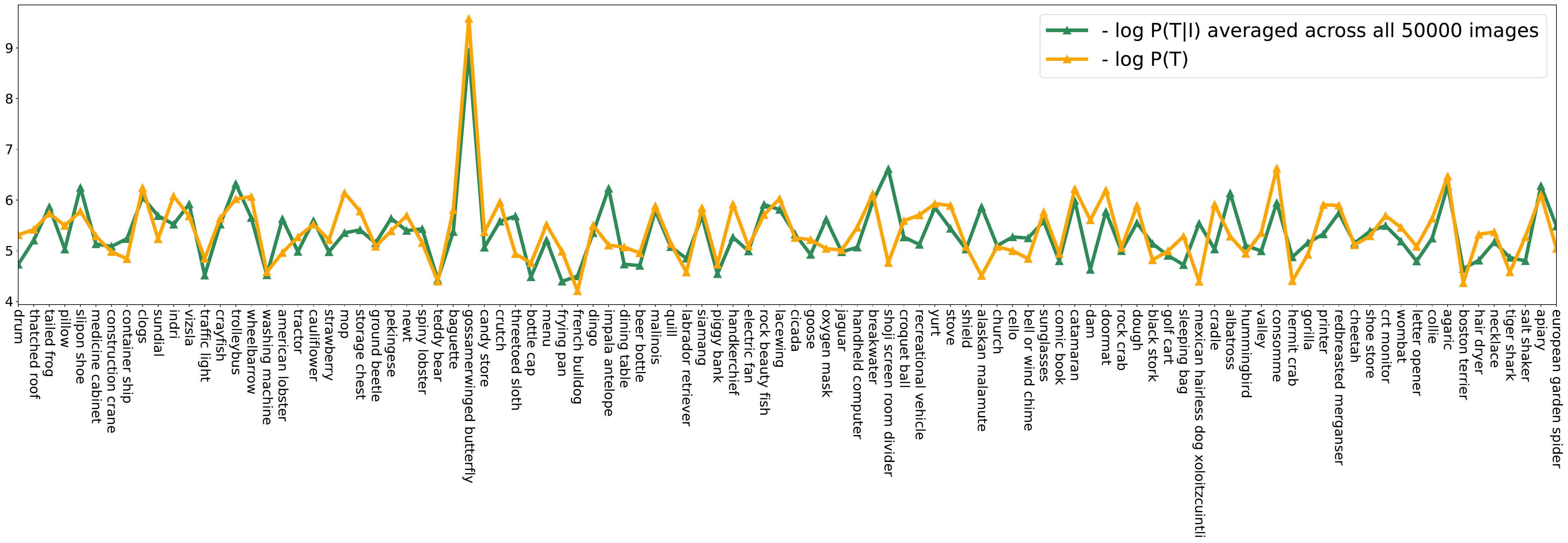}
 \end{tabular}
 \captionof{figure}{See correlations between the \textcolor{OliveGreen}{\bf green line} ($- \log P(T|I)$) and the \textcolor{YellowOrange}{\bf orange line} ($- \log P(T)$) on zero-shot ImageNet classification. $\log P(T|I)$ is predicted by a multimodal captioner trained on the Laion-5B dataset. $\log P(T)$ is predicted by an unimodal language model trained on the Laion-5B captions only.
 100 ImageNet classes are randomly sampled due to the limited space. The numerical correlation measurements between $\log P(T|I)$ and $\log P(T)$ for the whole 1000 classes are shown in~\tabref{tab: pcc_pti_pt}.}
 \label{fig: corr}
\end{table*}

\subsection{Evaluating Captioners with the Information Gain (IG)}

To alleviate the problem of strong correlation between caption likelihood conditioned on the image and caption priors, 
we tun to the following evaluation objective:
\begin{equation}
    \argmax_{i} \left[ \log \frac{P(T_{i}|I)}{P(T_{i})} \right]
\label{equ: IG_eval}
\end{equation}
Intuitively, this log-ratio measures the likelihood gain of the text $T_i$ given the image $I$. It is called pointwise mutual information~\cite{church1990word}.
The expectation of it over all the texts $T$ is the information gain~\cite{quinlan1986induction}. In this paper, we refer to pointwise mutual information as information gain.

We measure the correlation of $\log P(T_{i})$ and $\left[\log P(T_{i}|I) - \log P(T_{i}) \right]$ in~\tabref{tab: pcc_pti_pt}. It shows that the correlation is significantly reduced. Using this objective, we observe the performance gain in our ablation studies in~\secref{sec: ablate_alpha}. However, we notice that this quantity is negatively correlated with the caption priors, which is expected due to the existence of term $- \log P(T_{i})$ in~\equref{equ: IG_eval}.

To deal with this negative correlation problem, we further propose the following evaluation objective:
\begin{equation}
    \argmax_{i} \left[ \log P(T_{i}|I) - \alpha \log P(T_{i}) \right]
\label{equ: alpha_IG_eval}
\end{equation}
where $\alpha \in [0, 1]$ adjusts the degree of the text prior removal. Our thorough ablation study on $\alpha$ can be found in~\secref{sec: ablate_alpha}.

\subsection{Training Information Gain Captioners}

\renewcommand{\colwidthA}{2.0cm}
\begin{table*}[t]
 \centering
 \small
 \setlength{\tabcolsep}{0.0pt}
 \begin{tabular}{c}
    \includegraphics[width=0.95\textwidth]{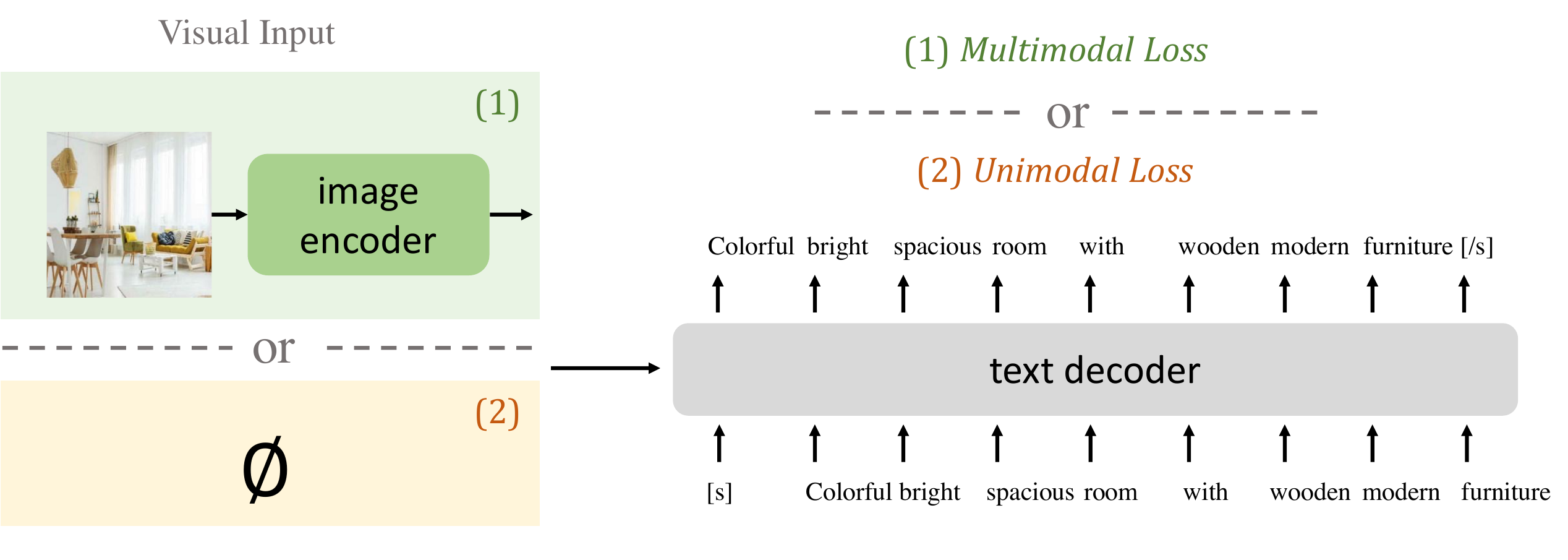}
 \end{tabular}
 \captionof{figure}{The training pipeline of IG captioner. IG captioner has two modes. Without any given images, IG captioner is an unimodal language model. When given an input image, IG captioner becomes the multimodal captioner that models the conditional probability of its caption. The image and caption are from the Laion-5B dataset.}
 \label{fig: training_pipeline}
\end{table*}

Without any changes on the training process, we observe the performance improvement of the captioner on the zero-shot classification task when using IG evaluation, which is shown in \secref{sec: IG_eval}. But in order to keep the training and evaluation consistent, we propose a two-objective approach to train the captioner, which is illustrated in~\figref{fig: training_pipeline}.

The first objective is the standard generative loss. Given an image, it supervises the captioner to predict the words of its caption in an autoregressive way:
\begin{equation}
    l_{\text{multimodal}} =\frac{1}{B} \sum_{i=1}^{B} \frac{1}{N^{i}} \sum_{n=1}^{N^{i}} \log \frac{1}{P(t_t^{i}|t_{1:n-1}^{i}, I^{i})}
\end{equation}
where $i$ is the index of a image-text pair $(I^{i}, T^{i})$ in one batch of size $B$. $N^{i}$ is the text token length.
Since this objective models the probability of a caption conditioned on an image, we call this objective the multimodal loss.

The second objective makes the captioner an autoregressive language model that is able to generate captions without given any image:
\begin{equation}
    l_{\text{unimodal}} =\frac{1}{B} \sum_{i=1}^{B} \frac{1}{N^{i}} \sum_{n=1}^{N^{i}} \log \frac{1}{P(t_n^{i}|t_{1:n-1}^{i})}
\end{equation}
We call this objective the unimodal loss.

Finally, we combine the above two losses in the following:
\begin{equation}
    L = \beta l_{\text{multimodal}} + \gamma l_{\text{unimodal}}
    \label{equ: IG captioner_training_obj}
\end{equation}
where $\beta$ and $\gamma$ are the weights for each loss. We perform an ablation study on the loss weights in~\secref{sec: ablate_loss_weights}. We name the captioner which is trained and evaluated with our proposed objectives IG captioner.

\section{Experiments}

\begin{table}[t]
  \small
  \centering
  \begin{tabular}{c|c|c|c}
    \toprule
    \multirow{1}{*}{model}  & \multirow{1}{*}{encoder}  & \multicolumn{1}{c|}{pretrain loss}    & \multirow{1}{*}{top-1 acc. (\%)} \\
    \midrule
    CLIP classifier  & ViT-B   & contrast & 63.8 \\
    captioner           & ViT-B   & generate  & 44.6 \\
    IG captioner        & ViT-B   & generate  & 64.3 \\
    \midrule
    CLIP classifier  & ViT-L   & contrast & 68.2 \\
    captioner           & ViT-L   & generate  & 51.7 \\
    IG captioner        & ViT-L   & generate  & 69.8 \\
    \bottomrule
  \end{tabular}
  \caption{Zero-shot ImageNet classification. All the models are pretrained with the public Laion-5B dataset. The image encoder, text decoder in captioner and IG captioner, and text encoder in CLIP classifier have the same layer number, and they are all built with standard transformer blocks.}
  \label{tab: zero_shot_imagenet}
\end{table}

\subsection{Zero-shot ImageNet classification}
\label{sec: main_imagenet}

\subsubsection{Settings.} We use the English subset of the public Laion-5B~\cite{schuhmann2022laion} dataset as the pretraining dataset for all experiments.
After pretraining, we perform zero-shot image classification on the ImageNet~\cite{deng2009imagenet} validation dataset. There are 1000 classes in this validation set and each class has 50 images. We build the captioner and IG captioner with an image encoder and a text decoder, and build the CLIP~\cite{radford2021learning} classifier with an image encoder and a text encoder. All the models consist of standard transformer blocks~\cite{vaswani2017attention}. We set the layer number of image encoder and text decoder / text encoder the same. We use a batch size 4096 for all the experiments. During evaluation, we follow CLIP to use 80 prompts for each class. We use the voting mechanism to combine the predictions from different prompts for the captioner and IG captioner. All the training and evaluation procedures are the same to ensure fair comparisons.

\subsubsection{Results.} The results are summarized in~\tabref{tab: zero_shot_imagenet}. First, compared with standard captioners, IG captioner achieves 19.7\% and 18.1\% top-1 accuracy improvements when using ViT-B and ViT-L~\cite{dosovitskiy2020image} as the image encoders. Second, compared with the strong discriminative baselines, CLIP classifiers, IG captioner achieves 0.5\% / 1.6\% top-1 accuracy improvements when using the image encoder ViT-B / ViT-L. These improvements demonstrate that IG captioner is a strong zero-shot classifier as a pure generative model.
We measure the correlations of the evaluation objective and the caption prior in~\secref{sec: ablate_alpha} to validate the effectiveness of the design of IG captioner. Besides, we perform the ablation studies in~\secref{sec: IG_eval} to demonstrate the efficacy of the IG evaluation on the standard captioner.

\subsection{Zero-shot Image-text Retrieval on MSCOCO and Flickr30K}

\subsubsection{Settings.} We use the same models in~\secref{sec: main_imagenet} pretrined on Laion-5B to perform zero-shot image-text retrieval on MSCOCO 5K and Flickr30K 1K test sets. There are 5 captions for each image. In order to measure the recalls, we compute a similarity matrix $\mathbf{M}_{a \times b}$ where $a$ and $b$ are the number of images and captions. For the CLIP classifier, $\mathbf{M}_{i, j}^{} = \text{E}_{I_{i}} \cdot \text{E}_{T_{j}}$, where $\text{E}_{T_{j}}$ and $\text{E}_{I_{i}}$ are the embeddings of the $i^{\text{th}}$ image and $j^{\text{th}}$ caption. For the captioner, we compute $\mathbf{M}_{i, j}^{} = \log P(T_{j}|I_{i})$ while for the IG captioner, we compute $\mathbf{M}_{i, j}^{} = \log P(T_{j}|I_{i}) - \log P(T_{j})$. 

\subsubsection{Results on MSCOCO.} We summarize the results on MSCOCO in~\tabref{tab: zero_shot_mscoco}. For the image $\rightarrow$ text retrieval, IG captioner significantly boosts the performance of a generative captioner.
IG captioner shows 22.7\% / 25.5\% / 22.8\% improvements over the captioner on R@1 / R@5 and R@10 with the ViT-B image encoder, and shows 22.9\% / 24.1\% / 20.3\% imporvements with the ViT-L image encoder. 
Furthermore, IG captioner achieves comparable performances with the CLIP classifier.
The performance gaps between the IG captioner and the CLIP classifier are 8.9\% / 6.6\% / 5.0\% on R@1 / R@5 / R@10 with the ViT-B encoder. While with the ViT-L image encoder, the gaps are further decreased to 6.5\% / 5.1\% / 3.5\% on R@1 / R@5 / R@10. 

Interestingly, for the text $\rightarrow$ image retrieval, we find that both the captioner and IG captioner have better performances than the CLIP classifier. Since our approach is to remove the bias in the text prediction, it does not affect the text $\rightarrow$ image retrieval. We leave this for the future research.

\begin{table}[t]
\centering
\begin{tabular}{@{}ccc|cccccccccccc@{}}
    \toprule 
    {} & {} & {} & \multicolumn{6}{c}{MSCOCO (5K test set)} \\
    {} & {} & {} & \multicolumn{3}{c}{image $\rightarrow$ text (\%)} & \multicolumn{3}{c}{text $\rightarrow$ image (\%)} \\
    \cmidrule(lr){4-6} \cmidrule(lr){7-9}
    model & encoder & pretrain loss & R@1 & R@5 & R@10 & R@1 & R@5 & R@10 \\
    \midrule
    CLIP classifier  & ViT-B & contrast  & 56.3 & 80.3 & 87.5 & 38.5 & 64.7 & 74.6 \\
    captioner           & ViT-B & generate  & 24.7 & 48.2 & 59.7 & 45.1 & 68.9 & 77.5 \\
    IG captioner        & ViT-B & generate  & 47.4 & 73.7 & 82.5 & 42.9 & 67.4 & 76.1 \\
    \midrule
    CLIP classifier  & ViT-L & contrast  & 59.3 & 83.0 & 89.5 & 42.1 & 67.9 & 77.4 \\
    captioner           & ViT-L & generate  & 29.9 & 53.8 & 65.7 & 49.2 & 72.8 & 80.8 \\
    IG captioner        & ViT-L & generate  & 52.8 & 77.9 & 86.0 & 48.5 & 71.6 & 80.2 \\
    \bottomrule
\end{tabular}
\caption{Zero-shot image-text retrieval on MSCOCO~\cite{chen2015microsoft}dataset. All the models are pretrained on Laion-5B dataset.}
\label{tab: zero_shot_mscoco}
\end{table}

\begin{table}[t]
\centering
\begin{tabular}{@{}ccc|cccccccccccc@{}}
    \toprule 
    {} & {} & {} & \multicolumn{6}{c}{Flickr30K (1K test set)} \\
    {} & {} & {} & \multicolumn{3}{c}{image $\rightarrow$ text (\%)} & \multicolumn{3}{c}{text $\rightarrow$ image (\%)} \\
    \cmidrule(lr){4-6} \cmidrule(lr){7-9}
    model & encoder & pretrain loss & R@1 & R@5 & R@10 & R@1 & R@5 & R@10 \\
    \midrule
    CLIP classifier  & ViT-B & contrast  & 85.2 & 97.2 & 98.7 & 68.2 & 88.9 & 93.6 \\
    captioner           & ViT-B & generate  & 46.9 & 73.1 & 81.5 & 72.2 & 90.8 & 94.3 \\
    IG captioner        & ViT-B & generate  & 60.6 & 90.0 & 94.7 & 70.0 & 89.3 & 93.6 \\
    \midrule
    CLIP classifier  & ViT-L & contrast  & 88.0 & 98.3 & 99.0 & 71.1 & 90.8 & 94.7 \\
    captioner           & ViT-L & generate  & 56.6 & 82.0 & 88.1 & 77.9 & 93.1 & 96.1 \\
    IG captioner        & ViT-L & generate  & 64.7 & 93.4 & 97.7 & 76.2 & 92.6 & 95.6 \\
    \bottomrule
\end{tabular}
\caption{Zero-shot image-text retrieval on Flickr30K~\cite{plummer2015flickr30k} dataset. All the models are pretrained on Laion-5B dataset.}
\label{tab: zero_shot_flickr}
\end{table}

\subsubsection{Results on Flickr30K.} The results are summarized in~\tabref{tab: zero_shot_mscoco}. We observe the similar phenomena on MSCOCO. For the image $\rightarrow$ text retrieval, on the one hand, IG captioner shows significant improvements over the captioner. With the ViT-B image encoder, IG captioner improves the R@1 / R@5 / R@10 by 13.7\% / 16.9\% / 13.2\%, and with the ViT-L image encoder, the improvements are 8.1\% / 11.4\% / 9.6\%. On the other hand, the performance gaps between the IG captioner and the CLIP classifier are significantly reduced. For R@1, the gaps are 24.6\% / 23.3\% with ViT-B / ViT-L image encoder. However, for R@5 / R@10, the gaps become 7.2\% / 4.0\% with ViT-B encoder, and 4.9\% / 1.3\% with ViT-L encoder. For the text $\rightarrow$ image retrieval, we observe the similar superiority of both the captioner and IG captioner over the CLIP classifier.

\section{Ablation Studies}

\subsection{Weights of the Text Prior Removal and Correlation Measurements}
\label{sec: ablate_alpha}

\begin{table*}[t]
  \centering
  \resizebox{1.0\textwidth}{!}{%
  \begin{tabular}{c|cc|cccc|cccc}
    \toprule
    {} & {} & {} & \multicolumn{4}{c|}{MSCOCO} & \multicolumn{4}{c}{Flickr30K} \\
    {} & \multicolumn{2}{c|}{ImageNet-1K} & \multicolumn{4}{c|}{image $\rightarrow$ text (\%)} & \multicolumn{4}{c}{image $\rightarrow$ text (\%)} \\
    \cmidrule(lr){2-3} \cmidrule(lr){4-7} \cmidrule(lr){8-11}
    \multirow{1}{*}{$\alpha$} & top-1 acc. (\%) & $r_{\log P(T), O_{\text{IG}}}$ & R@1 & R@5 & R@10 & $r_{\log P(T), O_{\text{IG}}}$ & R@1 & R@5 & R@10 & $r_{\log P(T), O_{\text{IG}}}$ \\
    \midrule
    0.0 & 45.4 & 0.83 & 22.0 & 44.3 & 56.2 & 0.87 & 42.2 & 68.4 & 76.8 & 0.90 \\
    0.1 & 49.1 & 0.80 & 24.9 & 48.3 & 60.9 & 0.85 & 45.6 & 71.9 & 81.3 & 0.88 \\
    0.2 & 52.5 & 0.76 & 28.8 & 53.2 & 65.0 & 0.81 & 50.2 & 76.8 & 84.7 & 0.85 \\
    0.3 & 55.6 & 0.70 & 33.0 & 58.2 & 68.4 & 0.77 & 54.5 & 80.0 & 87.7 & 0.81 \\
    0.4 & 58.4 & 0.62 & 36.7 & 61.7 & 72.6 & 0.71 & 58.2 & 84.2 & 90.3 & 0.76 \\
    0.5 & 60.7 & 0.52 & 40.2 & 65.7 & 75.9 & 0.62 & 61.6 & 86.4 & 92.3 & 0.70 \\
    0.6 & 62.7 & 0.40 & 43.2 & 69.4 & 78.8 & 0.51 & 63.2 & 88.2 & 93.6 & 0.61 \\
    0.7 & 63.8 & 0.25 & 46.2 & 72.0 & 80.9 & 0.36 & \cellcolor{SpringGreen}{63.4} & 89.0 & 94.0 & 0.48 \\
    0.8 & \cellcolor{SpringGreen}{64.3} & \cellcolor{SpringGreen}{0.08} & \cellcolor{SpringGreen}{47.4} & \cellcolor{SpringGreen}{73.7} & 82.5 & 0.19 & 60.6 & \cellcolor{SpringGreen}{90.0} & 94.7 & 0.32 \\
    0.9 & 63.9 & -0.10 & 46.4 & 73.2 & \cellcolor{SpringGreen}{82.5} & \cellcolor{SpringGreen}{0.0} & 56.1 & 89.7 & \cellcolor{SpringGreen}{95.8} & \cellcolor{SpringGreen}{0.12} \\
    1.0 & 62.9 & -0.27 & 43.3 & 70.6 & 81.4 & -0.21 & 50.5 & 86.7 & 94.5 & -0.09 \\
    \bottomrule
  \end{tabular}
  }
  \caption{Ablations on the $\alpha$ of the IG captioner evaluation objective, $O_{\text{IG}} = \log P(T|I) - \alpha \log P(T)$. All the performances are reported on zero-shot tasks. The IG captioner is pretrained on the Laion-5B dataset with the ViT-B image encoder. $r$ is the Pearson Correlation Coefficient (PCC). The text $\rightarrow$ image recalls for the retrieval tasks are not reported because the value of $\alpha$ does affect the image $\rightarrow$ text predictions. The best accuracies, recalls and PCCs are marked with the color SpringGreen.}
  \label{tab: ablate_alpha}
\end{table*}

\subsubsection{Settings.} In this section, we ablate the $\alpha$ in the evaluation objective of IG captioner, $\log P(T|I) - \alpha \log P(T)$. We use the same training and evaluation settings as the main experiments. We pretrain the IG captioner with the ViT-B encoder on the Laion-5B dataset and perform zero-shot ImageNet classification and zero-shot image-text retrieval on MSCOCO and Flickr30K.
We ablation the $\alpha$ from $0.0$ to $1.0$ with a step size $0.1$. For the retrieval tasks, we do not list the text $\rightarrow$ image recalls because the value of $\alpha$ does not affect the predictions of text $\rightarrow$ image retrieval. 
In order to validate our goal of removing the bias from text priors, we report the Pearson Correlation Coefficient (PCC) of $\log P(T)$ and the evaluation objective when varying the value of $\alpha$.

\subsubsection{Results.} We summarize the results in~\tabref{tab: ablate_alpha}. On the one hand, the smaller correlation between $\log P(T|I)$ and $\log P(T)$ generally corresponds to better performances. For the ImageNet classification, when $\alpha = 0.08$, the best PCC 0.08 and performance 64.3\% are achieved at the same time. For the retrieval tasks on MSCOCO / Flickr30K, the best PCC, 0.0 / 0.12, and the highest R@10, 82.5\% / 95.8\%, are obtained simultaneously with $\alpha = 0.9$. On the other hand, it requires a degree of text priors to achieve the best fine-grained performances on specific datasets. The highest R@1, 47.4\% / 63.4\%, correspond to the PCC, 0.19 / 0.48, on MSCOCO / Flickr30K.
$\alpha = 0.0$ corresponds to the bad strong correlations, resulting in the poor performances.
$\alpha = 1.0$ causes the negative correlations and produces the unsatisfying performances.
According to the ImageNet classification results with ViT-B encoder, we set $\alpha = 0.8$ for all experiments in this paper. 
For the ImageNet classification with ViT-L encoder and all the image-text retrieval tasks, we do not change this $\alpha$.

\subsection{Language Model + Captioner versus IG Captioner}

\subsubsection{Settings.} IG captioner has a text decoder that is able to predict both $\log P(T|I)$ and $\log P(T)$. One variant of IG captioner consists of a pair of models. One of them is the unimodal Language Model (LM) to predict $\log P(T)$. The other one is the multimodal captioner, a standard captioner to predict $\log P(T|I)$.
During inference, the prediction is based on the subtraction of their outputs. We call this LM + Cap approach.

\subsubsection{Results.} We report the results in~\tabref{tab: ablate_two_captioner}. It is observed that the LM + Cap approach performs worse than the IG captioner. Besides, it is less efficient due to the extra language model.
From another perspective, it demonstrates that the text decoder of IG captioner is able to perform both the multimodal and unimodal tasks well.

\begin{table}[t]
  \centering
  \begin{tabular}{ccc|c}
    \toprule
    \multirow{1}{*}{approach}  & \multirow{1}{*}{encoder} & \multirow{1}{*}{eval obj $\uparrow$} & \multirow{1}{*}{top-1 acc. (\%)} \\
    \midrule
    LM + Cap       & ViT-B  & $\log {P_{\theta}(T|I)} - \log {P_{\psi}(T)}$ & 64.0 \\
    IG captioner        & ViT-B  & $\log {P_{\theta}(T|I)} - \log {P_{\theta}(T)}$ & 64.3 \\
    \midrule
    LM + Cap       & ViT-L  & $\log {P_{\theta}(T|I)} - \log {P_{\psi}(T)}$ & 68.9 \\
    IG captioner        & ViT-L  & $\log {P_{\theta}(T|I)} - \log {P_{\theta}(T)}$ & 69.8 \\
    \bottomrule
  \end{tabular}
  \caption{Approach ablations on zero-shot ImageNet classification. All the models are trained on the Laion-5B dataset. LM + Cap stands for the approach that uses an unimodal Language Model (LM) and a standard multimodal captioner (Cap). The LM is trained with the captions only. $\psi$ represents the parameters of the LM while $\theta$ the parameters of the Cap.
  }
  \label{tab: ablate_two_captioner}
\end{table}

\renewcommand{\colwidthA}{2.0cm}
\begin{table}[t]
 \centering
 \setlength{\tabcolsep}{0.0pt}
 \begin{tabular}{c}
    \includegraphics[width=1.0\textwidth]{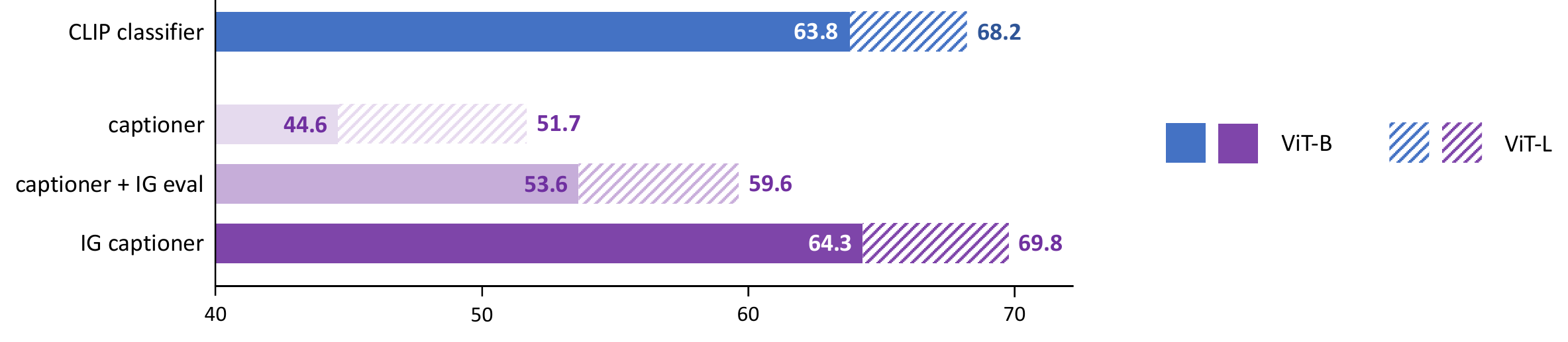}
 \end{tabular}
 \captionof{figure}{Ablations of the Information Gain (IG) evaluation on zero-shot ImageNet classification. All the models are trained on the Laion-5B dataset. Captioner + IG eval uses the evaluation objective, $\log P(T|I) - \alpha \log P(T|\mathbf{0})$.
 $\log P(T|\mathbf{0})$ is the prediction of the captioner with the input being the zero-intensity image. It is used to approximate $\log P(T)$, because the standard captioner after training is not able to directly predict $\log P(T)$.}
 \label{fig: ablate_ig_eval}
\end{table}

\begin{table}[t]
  \small
  \centering
  \begin{tabular}{c|c|c|c|c}
    \toprule
    generative classifier & model type & pretrain data & params & top-1 acc. (\%) \\
    \midrule
    Imagen~\cite{clark2023text,saharia2022palette} & text-to-image & L5B + Pty  & 2B    & 62.7 \\
    Stable Diffusion 2.0~\cite{li2023your,rombach2022high} & text-to-image & L5B  & 1.45B & 61.4 \\
    IG captioner              & image-to-text & L5B  & 1.07B  & 69.8 \\
    \bottomrule
  \end{tabular}
  \caption{Comparisons of generative classifiers on zero-shot ImageNet classification. L5B and Pty represent the Laion-5B dataset and a proprietary dataset.}
  \label{tab: compare_image_to_text}
\end{table}

\subsection{IG Evaluation}
\label{sec: IG_eval}

\subsubsection{Settings.} IG captioner differs from a standard captioner in both the training and evaluation phases. In this section, we study whether the IG evaluation will improve a standard captioner as a classifier. The IG evaluation requires the prediction of $\log P(T)$. Since the standard captioner can not directly predict this, we feed the captioner zero-intensity images to compute $\log P(T|\mathbf{0})$ as the approximation.

\subsubsection{Results.} We show the results in~\tabref{fig: ablate_ig_eval}. The IG evaluation is able to significantly improve the performance of the captioners. Specifically, with the ViT-B / ViT-L image encoder, IG evaluation increases the zero-shot top-1 ImageNet classification accuracy by $9.0\%$ / $7.9\%$. 
Adding the training part of IG captioner, the performances will be further improved by $10.7\%$ / $10.2\%$.

\subsection{Comparisons to Text-to-image Generative Classifiers}

\subsubsection{Settings.} There are two kinds of vision-language generative models: text-to-image generative models and image-to-text captioners. There have been recent works~\cite{li2023your,clark2023text} that successfully convert the large text-to-image diffusion models to the zero-shot generative classifiers. We compare IG captioner with these models.

\subsubsection{Results.} The results are summarized in~\tabref{tab: compare_image_to_text}. With the same or smaller amount of pretraining data, IG captioner demonstrates the best performances. Compared with the Text-to-Image Classifiers~\cite{clark2023text} and Diffusion Classifier~\cite{li2023your} using Imagen~\cite{saharia2022palette} and Stable Diffusion~\cite{rombach2022high}, respectively, IG captioner shows $7.1\%$ / $8.4\%$ zero-shot top-1 accuracy improvements on ImageNet. Besides, IG captioner is a more compact model and only uses $53.5\%$ / $73.8\%$ of their parameters.

\subsection{Weights of the Multimodal and Unimodal Training Losses}
\label{sec: ablate_loss_weights}

\begin{table}[t]
\centering
\begin{minipage}[t]{0.45\linewidth} %
\begin{tabular}{ccc}
    \toprule
    \multicolumn{2}{c}{training loss weights} & \multirow{2}{*}{top-1 acc. (\%)} \\
    \cmidrule(lr){1-2}
    \multirow{1}{*}{$\beta: \log P(T|I)$} & \multirow{1}{*}{$\gamma: \log P(T)$} & {} \\
    \midrule
    1.0 & 1.0 & 63.0 \\
    1.5 & 0.5 & 64.3 \\
    1.9 & 0.1 & 62.8 \\
    \bottomrule
  \end{tabular}
  \caption{Ablations of the training loss weights on zero-shot ImageNet classification.
  The performances are obtained from the IG captioners trained on the Laion-5B dataset with ViT-B encoder. $\beta$ and $\gamma$ are the weights of the multimodal and unimodal training objectives in~\equref{equ: IG captioner_training_obj}}
  \label{tab: ablate_loss_weights}
\end{minipage}
\hfill
\begin{minipage}[t]{0.45\linewidth} %
\centering
  \begin{tabular}{c|c}
    \toprule
    models & VQAv2 test-std acc. (\%) \\
    \midrule
    CLIP & N/A \\
    captioner & 74.5 \\
    IG captioner & 75.2 \\
    \bottomrule
  \end{tabular}
  \caption{Generative evaluation on VQAv2. All the models are pretrained on the Laion-5B dataset with ViT-L encoder and fine-tuned on the VQAv2 dataset. CLIP can not perform the VQA task as a discriminative model.}
  \label{tab: vqa_eval}
\end{minipage}
\end{table}

\subsubsection{Settings and Results.}
We ablate the multimodal loss weight $\beta$ and the unimodal loss weight $\gamma$ in the training objective of IG captioner (\equref{equ: IG captioner_training_obj}). 
We train the IG captioners with ViT-B encoder on the Laion-5B dataset and evaluate them on zero-shot ImageNet classification.
The results are shown in~\tabref{tab: ablate_loss_weights}. The equal weights do not produce the optimal results. 
We notice that image encoder is only trained with the multimodal loss while the text decoder is supervised by both the losses.
We set the multimodal loss weight $\beta = 1.5$ and the unimodal loss $\gamma = 0.5$ for all the other experiments.

\subsection{Generative Evaluation of IG Captioner}

\subsubsection{Settings and Results.}
We evaluate the generative capability of IG captioner in this section. Specifically, after pertaining on the Laion-5B dataset, we perform the generative fine-tuning on the VQAv2 dataset. The results are shown in~\tabref{tab: vqa_eval}. The IG captioner outperforms the captioner, showing that the information gain approach improves the captioner in both discriminative and generative tasks.

\section{Conclusion}

In this paper, we address the challenge of using a generative captioner for zero-shot classification tasks. We identify and mitigate the issue of linguistic priors dominating visual signals by introducing the Information Gain (IG) captioner with novel training and evaluation methods. Our results show that the IG captioner is an effective zero-shot classifier.

\subsubsection{Acknowledgements}
We gratefully acknowledge supports from ONR with N00014-23-1-2641.

\bibliographystyle{splncs04}
\bibliography{main}

\begin{thebibliography}{10}
\providecommand{\url}[1]{\texttt{#1}}
\providecommand{\urlprefix}{URL }
\providecommand{\doi}[1]{https://doi.org/#1}

\bibitem{achiam2023gpt}
Achiam, J., Adler, S., Agarwal, S., Ahmad, L., Akkaya, I., Aleman, F.L., Almeida, D., Altenschmidt, J., Altman, S., Anadkat, S., et~al.: Gpt-4 technical report. arXiv preprint arXiv:2303.08774  (2023)

\bibitem{alayrac2022flamingo}
Alayrac, J.B., Donahue, J., Luc, P., Miech, A., Barr, I., Hasson, Y., Lenc, K., Mensch, A., Millican, K., Reynolds, M., et~al.: Flamingo: a visual language model for few-shot learning. Advances in neural information processing systems  \textbf{35},  23716--23736 (2022)

\bibitem{bai2023qwen}
Bai, J., Bai, S., Yang, S., Wang, S., Tan, S., Wang, P., Lin, J., Zhou, C., Zhou, J.: Qwen-vl: A versatile vision-language model for understanding, localization, text reading, and beyond  (2023)

\bibitem{brown2020language}
Brown, T., Mann, B., Ryder, N., Subbiah, M., Kaplan, J.D., Dhariwal, P., Neelakantan, A., Shyam, P., Sastry, G., Askell, A., et~al.: Language models are few-shot learners. Advances in neural information processing systems  \textbf{33},  1877--1901 (2020)

\bibitem{chen2015microsoft}
Chen, X., Fang, H., Lin, T.Y., Vedantam, R., Gupta, S., Doll{\'a}r, P., Zitnick, C.L.: Microsoft coco captions: Data collection and evaluation server. arXiv preprint arXiv:1504.00325  (2015)

\bibitem{church1990word}
Church, K., Hanks, P.: Word association norms, mutual information, and lexicography. Computational linguistics  \textbf{16}(1),  22--29 (1990)

\bibitem{clark2023text}
Clark, K., Jaini, P.: Text-to-image diffusion models are zero-shot classifiers. arXiv preprint arXiv:2303.15233  (2023)

\bibitem{instructblip}
Dai, W., Li, J., Li, D., Tiong, A.M.H., Zhao, J., Wang, W., Li, B., Fung, P., Hoi, S.: Instructblip: Towards general-purpose vision-language models with instruction tuning (2023)

\bibitem{deng2009imagenet}
Deng, J., Dong, W., Socher, R., Li, L.J., Li, K., Fei-Fei, L.: Imagenet: A large-scale hierarchical image database. In: 2009 IEEE conference on computer vision and pattern recognition. pp. 248--255. Ieee (2009)

\bibitem{dosovitskiy2020image}
Dosovitskiy, A., Beyer, L., Kolesnikov, A., Weissenborn, D., Zhai, X., Unterthiner, T., Dehghani, M., Minderer, M., Heigold, G., Gelly, S., et~al.: An image is worth 16x16 words: Transformers for image recognition at scale. arXiv preprint arXiv:2010.11929  (2020)

\bibitem{gafni2022make}
Gafni, O., Polyak, A., Ashual, O., Sheynin, S., Parikh, D., Taigman, Y.: Make-a-scene: Scene-based text-to-image generation with human priors. In: European Conference on Computer Vision. pp. 89--106. Springer (2022)

\bibitem{ge2024visual}
Ge, Y., Zeng, X., Huffman, J.S., Lin, T.Y., Liu, M.Y., Cui, Y.: Visual fact checker: Enabling high-fidelity detailed caption generation. In: Proceedings of the IEEE/CVF Conference on Computer Vision and Pattern Recognition. pp. 14033--14042 (2024)

\bibitem{te2015basic}
te~Grotenhuis, M., Matthijssen, A.: Basic SPSS tutorial. Sage Publications (2015)

\bibitem{he2019pun}
He, H., Peng, N., Liang, P.: Pun generation with surprise. arXiv preprint arXiv:1904.06828  (2019)

\bibitem{hinton2007recognize}
Hinton, G.E.: To recognize shapes, first learn to generate images. Progress in brain research  \textbf{165},  535--547 (2007)

\bibitem{ho2020denoising}
Ho, J., Jain, A., Abbeel, P.: Denoising diffusion probabilistic models. Advances in neural information processing systems  \textbf{33},  6840--6851 (2020)

\bibitem{ho2022classifier}
Ho, J., Salimans, T.: Classifier-free diffusion guidance. arXiv preprint arXiv:2207.12598  (2022)

\bibitem{kosmos-1}
Huang, S., Dong, L., Wang, W., Hao, Y., Singhal, S., Ma, S., Lv, T., Cui, L., Mohammed, O.K., Liu, Q., Aggarwal, K., Chi, Z., Bjorck, J., Chaudhary, V., Som, S., Song, X., Wei, F.: Language is not all you need: Aligning perception with language models. ArXiv  \textbf{abs/2302.14045} (2023)

\bibitem{li2023your}
Li, A.C., Prabhudesai, M., Duggal, S., Brown, E., Pathak, D.: Your diffusion model is secretly a zero-shot classifier. arXiv preprint arXiv:2303.16203  (2023)

\bibitem{li2024llavanext-strong}
Li, B., Zhang, K., Zhang, H., Guo, D., Zhang, R., Li, F., Zhang, Y., Liu, Z., Li, C.: Llava-next: Stronger llms supercharge multimodal capabilities in the wild (May 2024), \url{https://llava-vl.github.io/blog/2024-05-10-llava-next-stronger-llms/}

\bibitem{li2015diversity}
Li, J., Galley, M., Brockett, C., Gao, J., Dolan, B.: A diversity-promoting objective function for neural conversation models. arXiv preprint arXiv:1510.03055  (2015)

\bibitem{li2023blip}
Li, J., Li, D., Savarese, S., Hoi, S.: Blip-2: Bootstrapping language-image pre-training with frozen image encoders and large language models. In: International conference on machine learning. pp. 19730--19742. PMLR (2023)

\bibitem{li2022blip}
Li, J., Li, D., Xiong, C., Hoi, S.: Blip: Bootstrapping language-image pre-training for unified vision-language understanding and generation. In: International conference on machine learning. pp. 12888--12900. PMLR (2022)

\bibitem{li2022contrastive}
Li, X.L., Holtzman, A., Fried, D., Liang, P., Eisner, J., Hashimoto, T., Zettlemoyer, L., Lewis, M.: Contrastive decoding: Open-ended text generation as optimization. arXiv preprint arXiv:2210.15097  (2022)

\bibitem{liu2021dexperts}
Liu, A., Sap, M., Lu, X., Swayamdipta, S., Bhagavatula, C., Smith, N.A., Choi, Y.: Dexperts: Decoding-time controlled text generation with experts and anti-experts. arXiv preprint arXiv:2105.03023  (2021)

\bibitem{liu2023improvedllava}
Liu, H., Li, C., Li, Y., Lee, Y.J.: Improved baselines with visual instruction tuning (2023)

\bibitem{liu2024llavanext}
Liu, H., Li, C., Li, Y., Li, B., Zhang, Y., Shen, S., Lee, Y.J.: Llava-next: Improved reasoning, ocr, and world knowledge (January 2024), \url{https://llava-vl.github.io/blog/2024-01-30-llava-next/}

\bibitem{liu2024visual}
Liu, H., Li, C., Wu, Q., Lee, Y.J.: Visual instruction tuning. Advances in neural information processing systems  \textbf{36} (2024)

\bibitem{ng2001discriminative}
Ng, A., Jordan, M.: On discriminative vs. generative classifiers: A comparison of logistic regression and naive bayes. Advances in neural information processing systems  \textbf{14} (2001)

\bibitem{nichol2021glide}
Nichol, A., Dhariwal, P., Ramesh, A., Shyam, P., Mishkin, P., McGrew, B., Sutskever, I., Chen, M.: Glide: Towards photorealistic image generation and editing with text-guided diffusion models. arXiv preprint arXiv:2112.10741  (2021)

\bibitem{OpenAI_GPT4_2023}
OpenAI: Gpt-4 technical report. ArXiv  \textbf{abs/2303.08774} (2023), \url{https://arxiv.org/abs/2303.08774}

\bibitem{pearson1895vii}
Pearson, K.: Vii. note on regression and inheritance in the case of two parents. proceedings of the royal society of London  \textbf{58}(347-352),  240--242 (1895)

\bibitem{peng2023kosmos}
Peng, Z., Wang, W., Dong, L., Hao, Y., Huang, S., Ma, S., Wei, F.: Kosmos-2: Grounding multimodal large language models to the world. arXiv preprint arXiv:2306.14824  (2023)

\bibitem{plummer2015flickr30k}
Plummer, B.A., Wang, L., Cervantes, C.M., Caicedo, J.C., Hockenmaier, J., Lazebnik, S.: Flickr30k entities: Collecting region-to-phrase correspondences for richer image-to-sentence models. In: Proceedings of the IEEE international conference on computer vision. pp. 2641--2649 (2015)

\bibitem{quinlan1986induction}
Quinlan, J.R.: Induction of decision trees. Machine learning  \textbf{1},  81--106 (1986)

\bibitem{radford2021learning}
Radford, A., Kim, J.W., Hallacy, C., Ramesh, A., Goh, G., Agarwal, S., Sastry, G., Askell, A., Mishkin, P., Clark, J., et~al.: Learning transferable visual models from natural language supervision. In: International conference on machine learning. pp. 8748--8763. PMLR (2021)

\bibitem{ramesh2022hierarchical}
Ramesh, A., Dhariwal, P., Nichol, A., Chu, C., Chen, M.: Hierarchical text-conditional image generation with clip latents, 2022. URL https://arxiv. org/abs/2204.06125  \textbf{7} (2022)

\bibitem{ranzato2011deep}
Ranzato, M., Susskind, J., Mnih, V., Hinton, G.: On deep generative models with applications to recognition. In: CVPR 2011. pp. 2857--2864. IEEE (2011)

\bibitem{reid2024gemini}
Reid, M., Savinov, N., Teplyashin, D., Lepikhin, D., Lillicrap, T., Alayrac, J.b., Soricut, R., Lazaridou, A., Firat, O., Schrittwieser, J., et~al.: Gemini 1.5: Unlocking multimodal understanding across millions of tokens of context. arXiv preprint arXiv:2403.05530  (2024)

\bibitem{rombach2022high}
Rombach, R., Blattmann, A., Lorenz, D., Esser, P., Ommer, B.: High-resolution image synthesis with latent diffusion models. In: Proceedings of the IEEE/CVF conference on computer vision and pattern recognition. pp. 10684--10695 (2022)

\bibitem{saharia2022palette}
Saharia, C., Chan, W., Chang, H., Lee, C., Ho, J., Salimans, T., Fleet, D., Norouzi, M.: Palette: Image-to-image diffusion models. In: ACM SIGGRAPH 2022 Conference Proceedings. pp. 1--10 (2022)

\bibitem{saharia2022photorealistic}
Saharia, C., Chan, W., Saxena, S., Li, L., Whang, J., Denton, E.L., Ghasemipour, K., Gontijo~Lopes, R., Karagol~Ayan, B., Salimans, T., et~al.: Photorealistic text-to-image diffusion models with deep language understanding. Advances in Neural Information Processing Systems  \textbf{35},  36479--36494 (2022)

\bibitem{schuhmann2022laion}
Schuhmann, C., Beaumont, R., Vencu, R., Gordon, C., Wightman, R., Cherti, M., Coombes, T., Katta, A., Mullis, C., Wortsman, M., et~al.: Laion-5b: An open large-scale dataset for training next generation image-text models. Advances in Neural Information Processing Systems  \textbf{35},  25278--25294 (2022)

\bibitem{sohl2015deep}
Sohl-Dickstein, J., Weiss, E., Maheswaranathan, N., Ganguli, S.: Deep unsupervised learning using nonequilibrium thermodynamics. In: International conference on machine learning. pp. 2256--2265. PMLR (2015)

\bibitem{su2022contrastive}
Su, Y., Lan, T., Wang, Y., Yogatama, D., Kong, L., Collier, N.: A contrastive framework for neural text generation. Advances in Neural Information Processing Systems  \textbf{35},  21548--21561 (2022)

\bibitem{team2023gemini}
Team, G., Anil, R., Borgeaud, S., Wu, Y., Alayrac, J.B., Yu, J., Soricut, R., Schalkwyk, J., Dai, A.M., Hauth, A., et~al.: Gemini: a family of highly capable multimodal models. arXiv preprint arXiv:2312.11805  (2023)

\bibitem{vaswani2017attention}
Vaswani, A., Shazeer, N., Parmar, N., Uszkoreit, J., Jones, L., Gomez, A.N., Kaiser, {\L}., Polosukhin, I.: Attention is all you need. Advances in neural information processing systems  \textbf{30} (2017)

\bibitem{wang2022ofa}
Wang, P., Yang, A., Men, R., Lin, J., Bai, S., Li, Z., Ma, J., Zhou, C., Zhou, J., Yang, H.: Ofa: Unifying architectures, tasks, and modalities through a simple sequence-to-sequence learning framework. In: International conference on machine learning. pp. 23318--23340. PMLR (2022)

\bibitem{yu2022coca}
Yu, J., Wang, Z., Vasudevan, V., Yeung, L., Seyedhosseini, M., Wu, Y.: Coca: Contrastive captioners are image-text foundation models. arXiv preprint arXiv:2205.01917  (2022)

\bibitem{yu2022scaling}
Yu, J., Xu, Y., Koh, J.Y., Luong, T., Baid, G., Wang, Z., Vasudevan, V., Ku, A., Yang, Y., Ayan, B.K., et~al.: Scaling autoregressive models for content-rich text-to-image generation. arXiv preprint arXiv:2206.10789  (2022)

\bibitem{zhu2023minigpt}
Zhu, D., Chen, J., Shen, X., Li, X., Elhoseiny, M.: Minigpt-4: Enhancing vision-language understanding with advanced large language models. arXiv preprint arXiv:2304.10592  (2023)

\end{thebibliography}
\end{document}